\documentclass[runningheads]{llncs}

\usepackage{eccv}

\usepackage{eccvabbrv}

\usepackage{graphicx}
\usepackage{booktabs}

\usepackage[accsupp]{axessibility}  %

\usepackage[pagebackref,breaklinks,colorlinks,citecolor=eccvblue]{hyperref}

\usepackage{orcidlink}

\usepackage[dvipsnames]{xcolor}
\usepackage{color}
\usepackage{booktabs}
\usepackage{makecell}
\usepackage{multirow}
\usepackage[export]{adjustbox}
\usepackage{tabu}
\usepackage{mathtools}
\usepackage{comment}
\usepackage{paralist}
\usepackage{tabularray}

\definecolor{turquoise}{cmyk}{0.65,0,0.1,0.3}
\definecolor{purple}{rgb}{0.65,0,0.65}
\definecolor{dark_green}{rgb}{0, 0.5, 0}
\definecolor{orange}{rgb}{0.8, 0.6, 0.2}
\definecolor{red}{rgb}{0.8, 0.2, 0.2}
\definecolor{darkred}{rgb}{0.6, 0.1, 0.05}
\definecolor{blueish}{rgb}{0.0, 0.3, .6}
\definecolor{light_gray}{rgb}{0.7, 0.7, .7}
\definecolor{pink}{rgb}{0.9, 0, 0.6}
\definecolor{greyblue}{rgb}{0.25, 0.25, 1}
\definecolor{teal}{rgb}{0.0, 0.4, 0.4}
\definecolor{chocolate}{rgb}{1.0, 0.4, 0.0}

\newcommand{\ky}[1]{#1}
\newcommand{\ws}[1]{#1}

\DeclareMathOperator*{\argmin}{arg\,min}

\newcommand{\point}{\mathbf{p}}
\newcommand{\points}{\mathbf{P}}

\newcommand{\initpoints}{\points_\text{in}}
\newcommand{\initpoint}{\point}

\newcommand{\voxsize}{V}
\newcommand{\voxelidx}{v}

\newcommand{\pixcolor}{\mathbf{C}}
\newcommand{\radiance}{\mathbf{c}}
\newcommand{\density}{\mathbf{\sigma}}
\newcommand{\query}{\mathbf{q}}
\newcommand{\ray}{\mathbf{r}}

\newcommand{\IE}{{\mathbb{E}}}

\newcommand{\level}{s}
\newcommand{\numlevel}{S}
\newcommand{\validlevels}{\mathcal{S}}

\newcommand{\voxel}{\mathbf{V}}
\newcommand{\features}{\mathbf{F}}
\newcommand{\feature}{\mathbf{f}}
\newcommand{\featmlp}{\mathcal{F}}

\newcommand{\nerfmlp}{\mathcal{M}}
\newcommand{\pointnerf}{\mathcal{E}}
\newcommand{\neighbor}{\mathcal{N}}
\newcommand{\threshold}{\tau}

\newcommand{\spam}{\mathcal{A}}
\newcommand{\render}{\mathcal{R}}
\newcommand{\volumerender}{\render}

\usepackage{amssymb}%
\usepackage{pifont}%
\newcommand{\cmark}{\ding{51}}%
\newcommand{\xmark}{\ding{55}}%
\newcommand{\greencheck}{{\color{green}\cmark}}

\newcommand{\redcross}{{\color{red}\xmark}}

\def \customparskip {0.3em}
\renewcommand{\paragraph}[1]{\vspace{\customparskip}\noindent\textbf{#1}}

\def\citet{\cite}

\begin{document}

\title{PointNeRF++: A multi-scale, point-based\\Neural Radiance Field}

\titlerunning{PointNeRF++}

\author{%
    Weiwei Sun$^{1}$ \quad 
    Eduard Trulls$^{2}$\quad
    Yang-Che Tseng$^{1}$ \quad
    Sneha Sambandam$^{1}$ \quad \\
    Gopal Sharma$^{1}$ \quad
    Andrea Tagliasacchi$^{3, 4, 5}$ \quad
    Kwang Moo Yi$^{1}$\quad \\ 
}

\authorrunning{Sun et al.}

\institute{ 
$^1$University of British Columbia \hspace{1pt}
$^2$Google Research\hspace{1pt} \\ 
\small
$^3$Google DeepMind\hspace{1pt}
$^4$Simon Fraser University \hspace{1pt}
$^5$University of Toronto \hspace{1pt} \\
\vspace{1pt}
\url{https://pointnerfpp.github.io}
}
\maketitle

\begin{abstract}
Point clouds offer an attractive source of information to complement images in neural scene representations, especially when few images are available.
Neural rendering methods based on point clouds do exist, but they do not perform well when the point cloud quality is low---\eg sparse or incomplete, which is often the case with real-world data.
We overcome these problems with a simple representation that aggregates point clouds at multiple scale levels with sparse voxel grids at different resolutions.
To deal with point cloud sparsity, we average across multiple scale levels---but only among those that are valid, \ie, that have enough neighboring points in proximity to the ray of a pixel.
To help model areas without points, we add a global voxel at the coarsest scale, thus unifying ``classical'' and point-based NeRF formulations.
We validate our method on the NeRF Synthetic, ScanNet, and KITTI-360 datasets, 
\ky{%
outperforming the state of the art, with a significant gap compared to other NeRF-based methods, especially on more challenging scenes.
}%

\end{abstract}

\begin{figure}[t]
\centering
\setlength{\tabcolsep}{1pt}
\renewcommand{\arraystretch}{0.2}
\includegraphics[width=\linewidth]{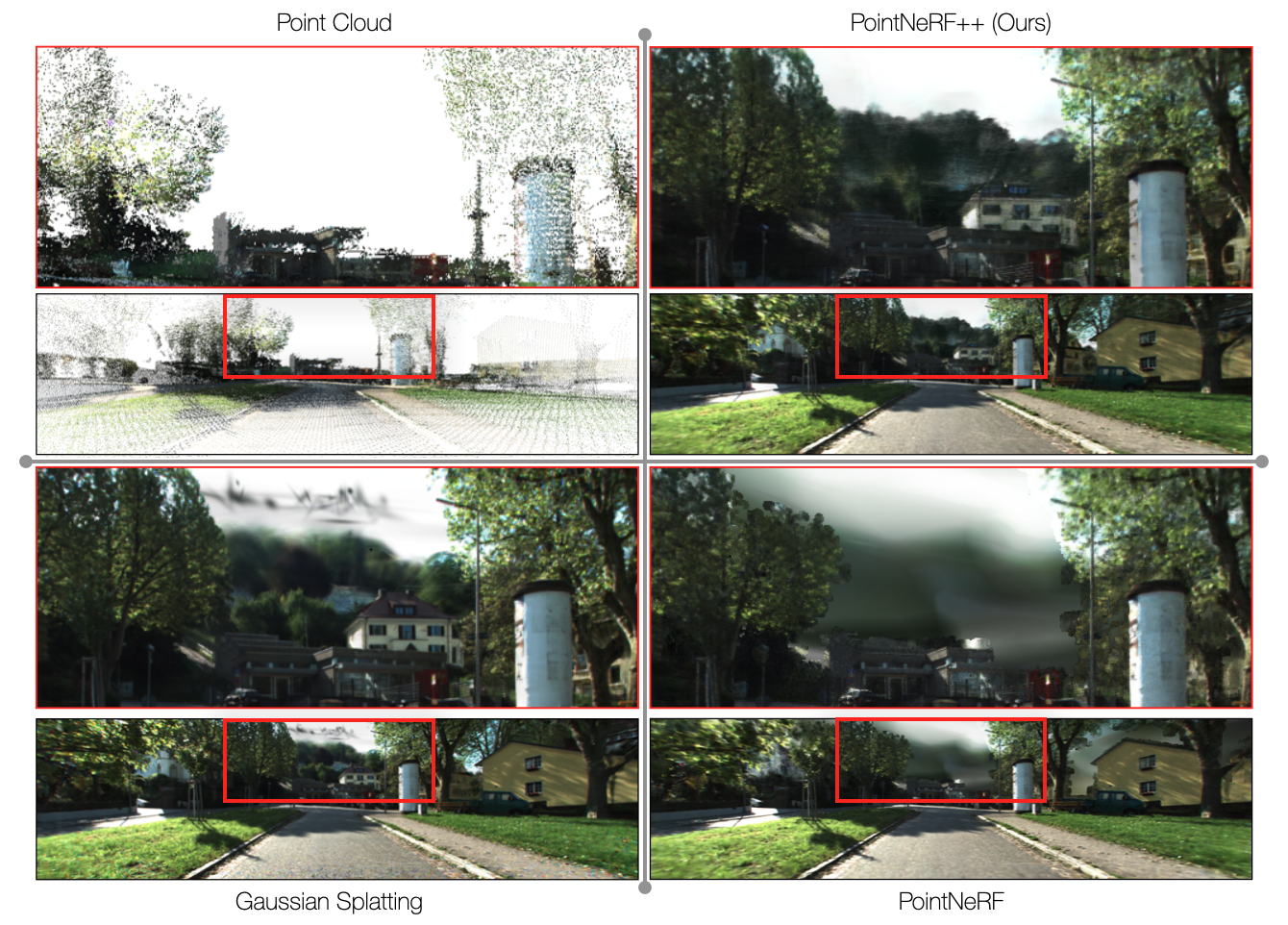}
\vspace{-1.7em}
\caption{
{\bf Teaser} -- 
We introduce a novel volume-rendering framework to effectively leverage point clouds for Neural Radiance Fields.
Our formulation aggregates points over multiple scales---including a global scale governing the entire scene, equivalent to the standard, point-agnostic NeRF. 
Our solution leads to much better novel-view synthesis in challenging real-world situations with sparse or incomplete point clouds.
\ky{Here, we show example renderings from the \texttt{KITTI-360} test set.}
\vspace{-1.7em}
}
\label{fig:teaser}
\end{figure}

\section{Introduction}
\label{sec:intro}
With the introduction of Neural Radiance Fields (NeRF)~\cite{mildenhall2020nerf}, the quality of novel-view synthesis from a collection of images has increased dramatically.
However, the problem is far from solved when \ky{field-of-view overlaps sparsely amongst cameras~\cite{yu2020pixelnerf, jain2021dietnerf, chen2021mvsnerf},}
which makes them difficult to apply to many uncontrolled, real-world scenarios. 
Researchers have attempted to solve this problem in various ways, including content-based regularization~\cite{jain2021dietnerf}, patch-based regularization~\cite{Niemeyer2021Regnerf}, image features~\cite{yu2020pixelnerf}, or diffusion priors~\cite{nerdi2023, wynn2023diffusionerf}.

One way to address this issue is to leverage point clouds obtained from additional sensors and/or photogrammetry~\cite{xu2022point, ost2022pointlightfields, rematas2022urban}.
The use of point clouds (as a representation) for neural rendering was pioneered by PointNeRF~\cite{xu2022point}, which demonstrated that point clouds can indeed help achieve higher-quality renderings.
However, as we demonstrate through experiments, the benefits of PointNeRF diminish when point clouds are sparse and/or incomplete.
This is often the case in most real-world applications, such as for point clouds obtained by LiDAR scanners in autonomous-driving datasets~\cite{Geiger2013IJRR, huang2019apolloscape, gated2depth2019, geyer2020a2d2, waymo2020, caesar2020nuscenes, liao2022kitti}.
We posit that this shortcoming is mainly due to a missing key element: the lack of \textit{multi-scale} modeling within the architecture of PointNeRF.
Multi-scale modeling is helpful in point cloud processing, as small `holes'~(regions without points) can often be naturally filled-in via multi-scale aggregation.
We liken this intuition to that followed by two seminal papers in point cloud semantic understanding---PointNet~\cite{qi2016pointnet} and PointNet++~\cite{qi2017pointnetplusplus}---where the latter improved upon the former by simply introducing a multi-scale network design, and the notion of hierarchical structure.

In this paper, we introduce a simple multi-scale representation for point cloud-based rendering.
Specifically, we aggregate point clouds at various scale levels, defined as voxel grids~(\cref{sec:multiscale}), up to a scale level that encompasses the \textit{entire} scene.
We then use this multi-scale representation to volume-render as in PointNeRF~(\cref{sec:volumerender})---but instead of averaging features locally, we do so across multiple scale levels.
This allows us to naturally deal with the sparsity of point clouds, without the need for failure-prone heuristics such as `pruning' and `growing'
from PointNeRF \cite{xu2022point}.
To account for the large support region required at coarser scales, we propose to replace the commonly used Multi-Layer Perceptron~(MLP) with a tri-plane representation~(\cref{sec:triplane}).
\ky{%
We note that using a single voxel at the coarsest scale (\ie,~global) is equivalent to a `standard' (\ie, not point-based) NeRF model.
Therefore, in a sense, our solution \textit{unifies} classical with point cloud-based NeRF formulations~(\cref{sec:multiscale}).
}%

As we illustrate in \Cref{fig:teaser}, our approach results in novel-view synthesis that is of significantly higher quality than previous methods. 
Compared to PointNeRF, our approach is able to deal with regions with both high and low point cloud density, even those without points (highlighted with red boxes).
\ky{The recently popular 3D Gaussian Splatting~\cite{kerbl3Dgaussians} also suffers at these empty regions as Gaussians are often initialized from point clouds.}
We evaluate \ky{our method} across three datasets, NeRF Synthetic~\cite{mildenhall2020nerf}, ScanNet~\cite{dai2017scannet}, and KITTI-360~\cite{liao2022kitti}, significantly outperforming the state of the art~(\cref{sec:results}).

We summarize our main contributions:
\begin{compactitem}
  \item we introduce an effective multi-scale representation for point-based NeRF;
  \item we propose to incorporate a global voxel/scale, uniting ``classical'' and point-based NeRF formulations;
  \item we propose to use a tri-plane representation for coarser scales to effectively cover larger support regions;
  \item we outperform all baselines, and specifically show large improvements over point-based \ky{NeRF}, especially when the point clouds are sparse or incomplete.
\end{compactitem}
\section{Related work}
\label{sec:related}
The introduction of Neural Radiance Fields~\cite{mildenhall2020nerf} represented a paradigm shift for scene representation and realistic novel-view synthesis.
NeRF employs a 5D implicit function to model a scene through a continuous volumetric approach, which estimates both density and radiance for any given position and direction.
Among many applications~\cite{gao2022nerf}, NeRFs have been used to reconstruct individual objects~\cite{mildenhall2020nerf} and unbounded scenes~\cite{mipnerf360}, in uncontrolled~\cite{martin2021nerf, chen2022hallucinated, xu2023neurallift} or dynamic environments~\cite{pumarola2020dynamicnerf, park2021nerfies, peng2021animatable, park2021hypernerf, jiang2022neuman}, in few-shot settings~\cite{yu2020pixelnerf, jain2021dietnerf, chen2021mvsnerf, Niemeyer2021Regnerf,wynn2023diffusionerf} and large urban landscapes~\cite{rematas2022urban, tancik2022block, wu2023scanerf}.

\paragraph{Accelerated training.}
While NeRF yields remarkable results, this comes at the cost of long training time, owing to the need to evaluate large MLP models hundreds of times for each pixel.
The prevailing approach to tackling this issue involves making a trade-off between compute and memory. 
This is achieved by storing features within various types of grids, including dense grids~\cite{yu_and_fridovichkeil2021plenoxels}, sparse grids~\cite{snerg2021}, multi-resolution hash grids ~\cite{mueller2022instant}, large sets of small MLPs~\cite{kilonerf}, low-rank tensor approximations of dense grids~\cite{Chen2022ECCV, kplanes_2023}, and hybrid planar and volumetric representations~\cite{reiser2023merf}.

\begin{figure}[t]
\centering
\includegraphics[width=\linewidth,trim={0 1.0cm 0 0},clip]{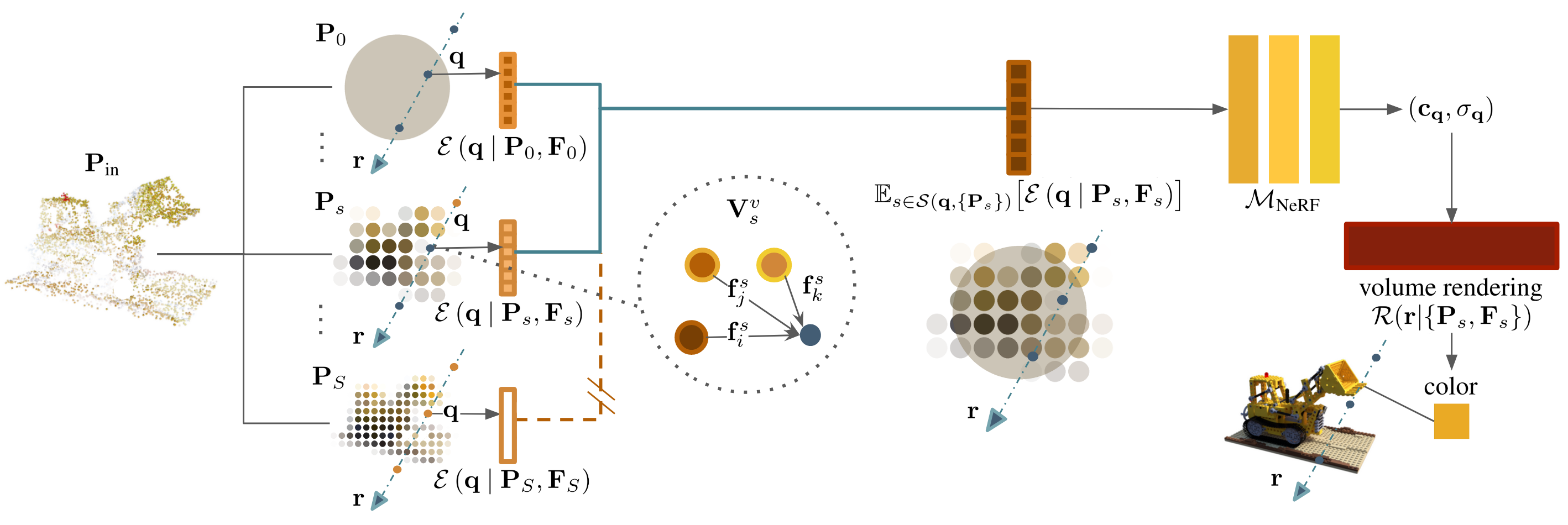}
\caption{%
{\bf Overview --}
Given an input point cloud, we aggregate it over multi-scale voxel grids~(\cref{sec:multiscale}).
\ky{For clarity, we draw the voxel grids in 2D.}
We then perform volume rendering based on points, relying on feature vectors stored thereon, which we aggregate across multiple scales~(\cref{sec:volumerender}).
Importantly, when aggregating across scales, we only take into account `valid' scales, \ie, those with nearby points---indicated with {\bf \textcolor{RoyalBlue}{solid blue lines}} \ky{and illustrated as the two overlaid scales in the middle}---naturally dealing with incomplete/sparse point clouds.
The coarsest scale (the top row in the figure) is a single, global voxel, equivalent to standard NeRF---\ie, it is not point-based.
}%
\label{fig:overview}
\vspace{-1.5em}
\end{figure}

\paragraph{Neural rendering with point clouds.}
While the techniques above can train efficiently, it is difficult to adapt them to model large environments.
An alternative approach is to use point clouds to model the geometric structure of the scene~\cite{xu2022point, ost2022pointlightfields, kerbl3Dgaussians, pointersect,chang2023neural}.
Point clouds can have variable density, helping allocate computational resources where needed, and conveniently (not) represent empty space.
To perform volume rendering, point cloud features are queried in the \textit{local} neighborhood of a ray to produce density and color.
These approaches can be classified on the basis of their neural point representations, \eg per-point features~\cite{xu2022point,chang2023neural}, factorized volumetric representations~\cite{trivol}, tetrahedral meshes~\cite{kulhanek2023tetranerf}, and learnable Gaussians~\cite{kerbl3Dgaussians}.

With PointNeRF, Xu et al.~\cite{xu2022point} and Chang et al.~\cite{chang2023neural} use point cloud data to learn per-scene representations, by querying per-point features within a local neighborhood.
Kulhanek et al. \citet{kulhanek2023tetranerf} create tetrahedra using the points from COLMAP~\cite{schonberger2016structure} and use barycentric interpolation to query the features within a tetrahedron.
Gaussian splatting~\cite{kerbl3Dgaussians} represents a 3D scene with 3D anisotropic Gaussians initialized by COLMAP, and optimizes their location to faithfully represent the scene.
\ky{%
Despite the high rendering quality, overall, Gaussian splatting is limited by the heuristics that they use to grow and prune points, similar to PointNeRF.
For example, as shown in \cref{fig:teaser}.
}%
In contrast to these works, our approach builds a hierarchy of feature representation, efficiently aggregating features in the local neighborhood at different levels; does not require optimizing the location of the points \ky{nor heuristics to grow and prune points}; and leads to superior performance even with sparse or incomplete point clouds.

Finally, rather than using geometric proximity, one can learn a point-to-query affinity function via transformers.
Ost et al. \citet{ost2022pointlightfields} use transformers to combine features of points along a ray to predict its color.
A shortcoming of this approach is that it does not take into account occlusions and combines all points in the neighborhood of a ray.
Similarly, Chang et al.~\citet{pointersect} use a set-transformer to find ray-surface intersections and use local features and blending weights to estimate ray colors.
Both of these approaches are different from ours, as we employ geometric, rather than learned, proximity.

\section{Method}
\label{sec:method}
\label{sec:overview}

An overview of our method is shown in \cref{fig:overview}.
We build a representation starting from an input point cloud, which we then use to volume-render~\cite{mildenhall2020nerf} a scene.
Specifically, given an input point cloud $\initpoints$, we spatially aggregate the point cloud to build a \textit{point cloud hierarchy} with $\numlevel$ levels.
Denoting this operation as $\spam(.)$, we write
\begin{equation}
    \{\points_\level\}_{\level=1}^\numlevel = \spam(\initpoints),
    \label{eq:voxelizehighlevel}
\end{equation}
and equip each point cloud level $\points_\level$ with randomly initialized point features $\features_\level$.
We then optimize the features $\features_\level$ by volume-rendering them along a ray (pixel) $\ray$ by~$\render(.)$, so that the estimated color matches that of the ground-truth pixels $\pixcolor_\text{gt}$, using a photometric loss:
\begin{equation}
    \argmin_{\{\features_\level\}} \:\: \IE_\ray
    \left[
    \|\pixcolor_\text{gt}(\ray) - \render(\ray | \{\points_\level, \features_\level\})\|_2^2
    \right] .
    \label{eq:objective}
\end{equation}
We next detail our multi-scale aggregation strategy to define a hierarchical representation for point clouds (\cref{sec:multiscale}), and how we use it to volume-render a scene (\cref{sec:volumerender}).
Finally, we propose to use a tri-plane-based feature representation in lieu of MLPs, in order to obtain a good trade-off between representation capacity and speed (\cref{sec:triplane}).

\subsection{Multi-scale aggregation\texorpdfstring{ -- $\spam$}{}}
\label{sec:multiscale}

We first detail our aggregation operation $\spam$ in \cref{eq:voxelizehighlevel}.
To obtain a point cloud that represents a desired scale level $\level$, we cluster based on voxels.
At level $\level$, consider a regular grid of resolution $\voxsize_\level{\times}\voxsize_\level{\times}\voxsize_\level$, consisting of a set of voxels $\{\voxel_\level^\voxelidx\}$.
We perform voxel-wise clustering to determine one representative point per voxel as
\begin{equation}
    \point_\level^\voxelidx = \mathbb{E}_{\initpoint \in \voxel_\level^\voxelidx}\left[\initpoint\right] 
    \quad\text{s.t.}\quad
    \initpoint \in \initpoints
    .
    \label{eq:voxelize}
\end{equation}

Importantly, note that this is performed only over non-empty voxels, hence the resulting representation is \textit{sparse}.
Note also that the aggregation is built at each scale level {\it independently}, and that while some fine-grained scales may not have valid aggregated points, more space regions will be covered at the coarser scales.
This allows for point clouds with \textit{variable density}, or even \textit{incomplete} ones to a certain degree, to be dealt with naturally.
Finally, we set the coarsest voxel to cover the entire scene, effectively setting $\point_0^0 = \mathbb{E}_{\initpoint \in \initpoints}\left[\initpoint\right]$.
This coarsest scale can also be understood as a global NeRF model that is 
independent of the local distribution of the point cloud---providing a unified representation for both standard and point-based NeRF.

\begin{figure}[t]
\centering
\includegraphics[width=.8\linewidth]{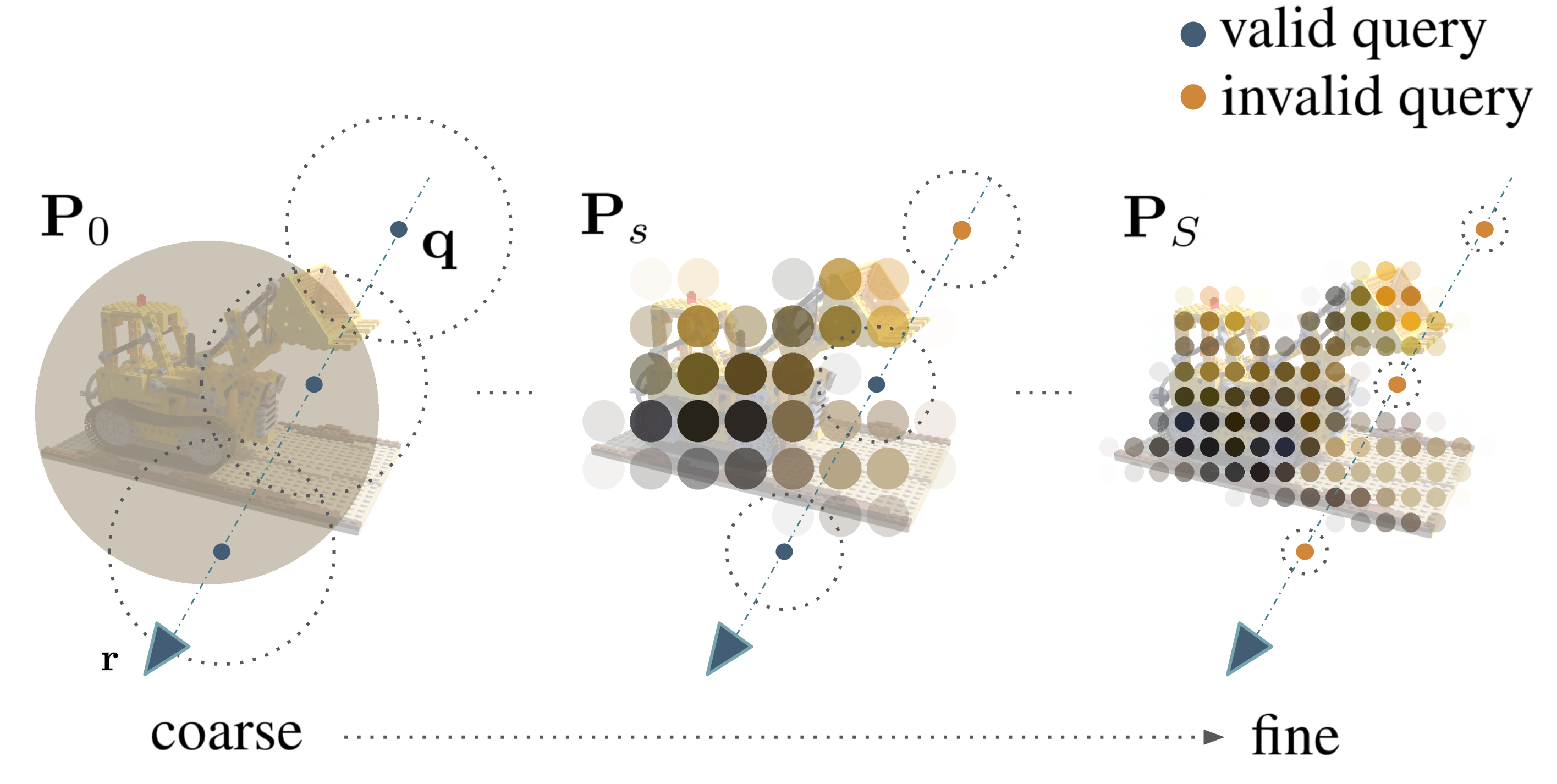}
\caption{{\bf Increasing coverage with multiple scales --}
We illustrate our sparse, hierarchical representation at three granularity levels, including a single, global voxel (left). We also show three query points, with their respective neighborhoods (dotted circles) at each scale level---color-coded in {\bf \textcolor{RoyalBlue}{blue}} if they have neighbouring features, and in {\bf \textcolor{Orange}{orange}} if they do not. Our multi-scale approach naturally fills in empty regions, removing the need for failure-prone region-growing heuristics \cite{xu2022point}.
Drawn in 2D, for clarity.}
\label{fig:scalelevel}
\vspace{-1em}
\end{figure}

\subsection{Point-based rendering\texorpdfstring{ -- $\volumerender$}{}}
\label{sec:volumerender}
We use volume rendering to render an image from the multi-scale point cloud.
Given a set of quadrature points along ray~$\query \in \ray$, let us denote the volume rendering integral~\cite{mildenhall2020nerf}
\begin{equation}
    \hat{\pixcolor}_\ray = 
    \volumerender_{\query \in \ray}\left(
        \radiance_\query, \density_\query
    \right)
    ,
\end{equation}
where $\radiance_\query$ is the radiance and $\density_\query$ is the density of a location~$\query$ in space.
To obtain these values, we operate on our point cloud hierarchy, as opposed to the raw point cloud $\initpoints$ as in PointNeRF~\cite{xu2022point}.
More explicitly, we extend~\cite{xu2022point} to multiple scales by averaging over valid scale levels, \ie, scale levels with any points within the vicinity of $\query$:
\begin{equation}
    \radiance_\query , \density_\query = \nerfmlp\Bigl(
    \IE_{\level \in \validlevels(\query, \{\points_\level\})}\bigl[
        \pointnerf\left(
            \query ~|~ \points_\level, \features_\level
        \right)
    \bigr]
    \Bigr)
    ,
    \label{eq:multiscalefeat}
\end{equation}
where $\validlevels(\query, \{\points_\level\})$ is the set of valid scale levels associated to query $\query$; $\pointnerf$ is the feature extraction operation in PointNeRF~\cite{xu2022point} that converts the point cloud into a feature embedding at the query location $\query$; and $\nerfmlp$
is an MLP that converts those feature embeddings into radiance and density.
We now describe $\validlevels$ and $\pointnerf$ in more detail.

\paragraph{Valid scale levels -- $\validlevels(\query, \{\points_\level\})$.}
Given a scale $\numlevel$, define $\neighbor$ the local neighbors of $\query$ within distance~$\threshold \voxsize_\level$, \ky{where $\threshold$ is the threshold ratio}:
\begin{equation}
    \neighbor(\query, \points_\level) = \bigl\{\point ~|~ 
    \point \in \points_\level \And
    \|\point - \query\|_2 \le \threshold\voxsize_\level \bigr\}
    .
    \label{eq:neighbor}
\end{equation}
which is then aggregated across levels to define:
\begin{equation}
    \validlevels(\query, \{\points_\level\})
    =
    \left\{
        \level \;|\; \neighbor(\query, \points_\level) \neq \varnothing
    \right\},
\end{equation}

\paragraph{Point cloud to feature embedding -- $\pointnerf(\query ~|~ \points_\level, \features_\level)$.}
We aggregate the features within the support defined by~\cref{eq:neighbor} using normalized inverse-distance weights~$w(\point, \query) = (\|\point - \query\|_2 + \varepsilon)^{-1}$, where $\varepsilon$ is a small number to avoid numerical problems:
\begin{equation}
    \pointnerf(\query | \points_\level, \features_\level)
    \!=\! 
    \frac{
    \sum_{\point \in \neighbor(\query, \points_\level)} 
    w(\point, \query) \:
    \featmlp\bigl(
        \feature_\point, \point-\query
    \bigr)     
    }{
    \sum_{\point \in \neighbor(\query, \points_\level)} 
    w(\point, \query)
    }
    ,
\end{equation}
where $\featmlp$ is a learnable function, and $\feature_\point$ is the feature in $\features_\level$ corresponding to $\point \in \points_\level$.
Note that this is a simplified version of PointNeRF~\cite{xu2022point}, as we do not use `per-point` weights~\cite[Sec.~4.1]{xu2022point}, which we experimentally found to not contribute to improvements in rendering quality.
{%
Rather than relying on large MLPs to implement $\featmlp$ at coarse levels $\level$, we employ a tri-plane representation, described in~\cref{sec:triplane}.
This effectively increases the representation power of $\featmlp$ at coarse levels so that less populated regions in space can still be modeled effectively, without incurring an excessive computational burden.
}%

\subsection{Per-point tri-plane features}
\label{sec:triplane}
As illustrated in \Cref{fig:scalelevel}, points in coarser levels represent larger regions, and thus require preserving more information into each point feature.
We could solve this by increasing either the feature dimension or the capacity of the MLPs used to parameterize~$\featmlp$.
They both come with a hefty price, greatly increasing the computational cost incurred to evaluate~$\featmlp$.
Instead, we build on recently-proposed factorized representations~\cite{kplanes_2023}, and represent local features with a local \textit{tri-plane} factorization.
In more detail, we store features within three orthogonal feature planes $\feature_\point \equiv \{\feature_\point^{XY}, \feature_\point^{YZ}, \feature_\point^{XZ} \}$, which are then accessed at (local) 3D coordinates \mbox{$\mathbf{u}=(\query-\point) / (\threshold \voxsize_\level)$}:
\begin{align}
\featmlp\bigl( \feature_\point, \mathbf{u} \bigr) =     
\feature_{\point}^{XY} [\mathbf{u}] +
\feature_{\point}^{YZ} [\mathbf{u}] + 
\feature_{\point}^{XZ} [\mathbf{u}]
,
\label{eq:triplane}
\end{align}
where $\feature_{\point}^{**}[\mathbf{u}]$ denotes querying the plane at position $\mathbf{u}$ with bilinear interpolation.
We combine tri-planes at coarser levels with the standard MLPs at finer levels, where we find the latter are sufficient (see \cref{sec:setup} for details).
At first glance, \cref{eq:triplane} may seem like a large deviation from using an MLP, since the features seem to be independent of each other, due to the lack of a \emph{shared} MLP.
Note, however, that those features are eventually processed by the shared decoder $\nerfmlp$ that converts them into radiance field values.
Finally, we note that at the coarsest scale level, \ie, the global voxel, our representation is effectively \mbox{K-Planes}~\cite{kplanes_2023}.

\section{Results}
\label{sec:results}

\subsection{Experimental setup}
\label{sec:setup}

\vspace{-\customparskip}
\paragraph{Datasets and metrics.}
We primarily use Peak Signal-to-Noise Ratio (PSNR) as a metric, and also structural (SSIM \cite{wang2004image}) and perceptual (LPIPS \cite{zhang2018unreasonable}) similarity. We evaluate our method on three well-known datasets:
\setdefaultleftmargin{0em}{0em}{}{}{}{}
\begin{compactitem}
   \item \texttt{KITTI-360}~\cite{liao2022kitti}
    is a recent benchmark of outdoor driving sequences, highly challenging due to the sparsity of views, which have much less visual overlap than other datasets.
    Each sequence consists of an average of 80 images.
    We use a random 10\% subset for validation purposes, and also for our ablation study, as the ground truth for the test set is not publicly available.
    To obtain results on the test set, we follow the standard practice of training with the entire training set, to roughly the same number of iterations required for convergence, discovered with the validation split.
    We use the point clouds provided with the dataset, from LiDAR scans that are accumulated over all views.
    As this accumulated point cloud is very dense, we resample it over a grid with a cell size of 8cm, and remove points outside the camera frustum of the training views to make it more tractable.

   \item \texttt{ScanNet}~\cite{dai2017scannet} 
    is a dataset of indoor scans.
    We use the point clouds provided with the dataset, which are sampled from mesh reconstructions using RGB-D cameras with BundleFusion~\cite{dai2017bundlefusion}.
    Following PointNeRF~\cite{xu2022point}, we evaluate on two scenes, Scene-101 and Scene-241.
    The point cloud in Scene-101 has more incomplete regions, which makes it harder than Scene-241.
    As in PointNeRF~\cite{xu2022point}, we sample 20\% of the images, \ie 1463 images for Scene-241, and 1000 images for Scene-101, for training, and use the rest for evaluation. We use the code provided by \cite{xu2022point}.

    \item \texttt{NeRF Synthetic}~\cite{mildenhall2020nerf} is a synthetic dataset with eight objects, each with 100 training images and 200 test images. 
    The images are purely synthetic, rendered with Blender.
    We use this dataset, as in PointNeRF~\cite{xu2022point}, to validate our method when the scene is favorable to the standard NeRF setting.
    We take the point clouds provided by PointNeRF~\cite{xu2022point}, which are obtained with COLMAP~\cite{schonberger2016structure}.
\end{compactitem}

\paragraph{Implementation details.}
We implement our method with PyTorch~\cite{paszke2017automatic}.
We use a total of 5 scale levels, including the global scale.
We use a tri-plane resolution of $512\times 512$ for the global scale level.
For the largest (\ie, coarsest) two of the remaining scale levels we use tri-planes, 
\ky{%
where each tri-plane is built as a small two-layer pyramid with $4\times 4$ and $2\times 2$ grid.
}%
For all tri-planes, we store 32-dimensional feature vectors followed by a four-layer MLP with 64 neurons.
For the remaining two (\ie, finest) scales, we simply use 32-dimensional point features and a four-layer MLP with 64 neurons. 
To further allow for the global scale to capture details that may be beyond the capacity of its resolution, we augment the features extracted from the global tri-plane with positional encodings with 5 frequencies, as in \cite{xu2022point}. 
We found this to be especially important when modeling large scenes, such as for \texttt{KITTI-360}.
For $\nerfmlp$, we use one linear layer for density prediction and a four-layer MLP with 64 neurons for its hidden layers for color prediction.    

To speed up neighborhood search, we rely on voxel-grid-based approximate nearest neighbors, as in \cite{xu2022point}.
We use the same search radius as our neighborhood threshold $\threshold$ in \cref{eq:neighbor} after normalizing so that the approximate search is equivalent to a ball query.
For speed-ups and to limit GPU memory growth, we set the maximum number of neighboring points to 8 for \texttt{ScanNet} and \texttt{NeRF Synthetic}, and 6 for \texttt{KITTI-360}, as the scenes are larger.
We follow PointNeRF~\cite{xu2022point} to sample 400 points for each ray on \texttt{ScanNet} and \texttt{NeRF Synthetic}. 
As \texttt{KITTI-360} is larger, we sample 1,000 points for each ray to compensate.
We use the contraction function of \citet{mipnerf360} for regions outside the bounding box of the point cloud.
For \texttt{KITTI-360}, as in \citet{rematas2022urban}, we model the sky using a four-layer MLP that maps ray direction to color.
\ky{To improve sample efficiency, we also use a proposal network~\cite{mipnerf360}.}

We train our model with a single NVidia V100 GPU for 200k iterations.
For the learning rate we follow \citet{xu2022point} and use the same exponential decaying scheduling with an initial learning rate of 5e-4 for $\nerfmlp$ and a larger initial learning rate of 2e-3 for $\features$.
Following ~\cite{xu2022point}, we decay every 1000k steps with a rate of 0.1.
We will make our code available with an Apache 2.0 license, for reproducibility.

\begin{figure}[t]
    \centering
    \includegraphics[width=\linewidth]{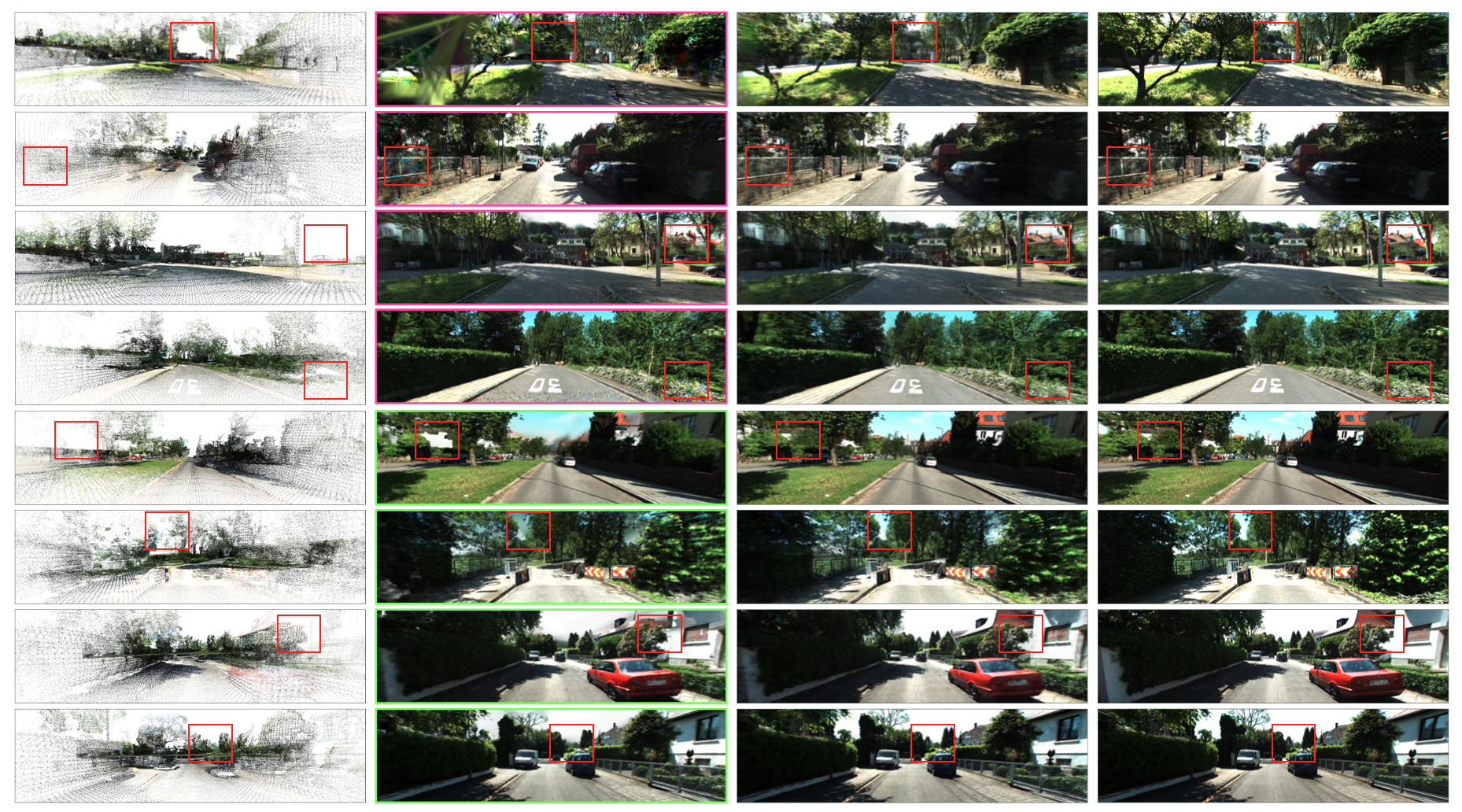}
    \renewcommand{\arraystretch}{0.3}
    \begin{tblr}{colspec={X[c]X[c]X[c]X[c]}}
    \scriptsize{(a) Input point cloud~\cite{dai2017scannet}} & \scriptsize{(b) \textcolor{pink}{Gauss. Splat.}~\cite{kerbl3Dgaussians} $\&$ \textcolor{green}{PointNeRF}~\cite{xu2022point}} & \scriptsize{(c) PointNeRF++ (Ours)} & \scriptsize{(d) Ground truth}\\
    \end{tblr}
    \caption{
    {\bf Examples on \texttt{KITTI-360} -- }
    We show novel-view renderings obtained with our method, 3D Gaussian splatting~\cite{kerbl3Dgaussians} (pink colored, 1-4 rows ) and PointNeRF~\cite{xu2022point} (green colored, 4-8 row) on a challenging outdoors dataset, using the same point clouds as input. 
    Our approach provides significantly sharper renderings with more details, and better coverage in areas without points, where Gaussian Splatting and PointNeRF produce highly salient artifacts highlighted with red boxes.
    }
    \label{fig:qualitative_kitti}
    \vspace{-3mm}
\end{figure}

\subsection{KITTI-360 results -- \cref{fig:qualitative_kitti} and \cref{tab:kitti}}
We first compare our method to the state of the art on \texttt{KITTI-360}~\cite{liao2022kitti}, a challenging outdoors dataset with incomplete point clouds from real LiDAR measurements.

\paragraph{Baselines.}
We report numbers on the hidden test set, which requires uploading samples to the evaluation server to compare with methods featured on the public leaderboard. 
We also report numbers on our validation split for methods that do not have an entry in the public leaderboard.
We consider methods based on images and, optionally, semantics~\cite{mildenhall2020nerf,barron2021mip,zhang2023nerflets,kundu2022panoptic,riegler2020free,wang2023planerf} as well as those that use the LiDAR point clouds~\cite{liao2022kitti,kerbl3Dgaussians,xu2022point,kopanas2021point}.
Note that for all PointNeRF experiments in this paper we use the point `pruning' and `growing' heuristics introduced in their work~\cite[Sec.~4.2]{xu2022point}, which aim at improving geometry modeling and image rendering quality, and can help deal with point cloud sparsity---our algorithm does not rely on it.

\paragraph{Discussions.}
\ky{%
We show qualitative highlights in \cref{fig:qualitative_kitti} and report results on \cref{tab:kitti}.
Our method achieves a new state of the art in the color-only category among NeRF methods, and performs on par with methods that also use semantic supervision and Gaussian Splatting.
}%
Importantly, we significantly improve over other point-based methods.
Compared with PointNeRF, our approach yields better renderings on regions where the point cloud is sparse, and the global scale allows us to tackle those with no nearby points, such as structures too far away to be captured by LiDAR.
\ky{%
Also of note is how Gaussian splatting, while it provides improved rendering quality in terms of SSIM and LPIPS thanks to its SSIM-based loss, can have failure modes, as highlighted in \cref{fig:qualitative_kitti}.
We thus believe combining our strategy of utilizing our multi-scale strategy with Gaussian splatting can further lead to performance improvements, but we leave as future work.
}%
Please refer to the appendix for video examples.

\begin{table}[t]
\caption{\textbf{Results on KITTI-360 \cite{liao2022kitti} --} Our method achieves the best performance among methods that supervise only with \textit{color}. 
It performs on par with those that also rely on \textit{semantics}.
We provide results on the (public) validation set and the (hidden) test set as some baselines have results for one, but not the other. 
}
\label{tab:kitti}
\resizebox{\linewidth}{!}{
\setlength{\tabcolsep}{8pt}
\begin{tabular}{llccccccc}
\toprule
                                    &&            & \multicolumn{3}{c}{{\bf Validation}} & \multicolumn{3}{c}{{\bf Test}} \\
                                                                 \cmidrule(lr){4-6} \cmidrule(lr){7-9}
                                    && \makecell{Uses \\points} & PSNR$\uparrow$            & SSIM$\uparrow$    & LPIPS$\downarrow$  & PSNR$\uparrow$      & SSIM$\uparrow$      & LPIPS$\downarrow$    \\
\midrule
\multirow{3}{*}{\rotatebox[origin=c]{90}{ {\scriptsize {w/ Sem.}} }}
& Nerflets~\cite{zhang2023nerflets}                             & \redcross               & --               & --        & --       & 21.69      & --          & --         \\
& PNF~\cite{kundu2022panoptic}& \redcross               & --               & --        & --       & \underline{22.07}      & \underline{0.820}& \underline{0.221}     \\
& PLANeRF~\cite{wang2023planerf}                             & \redcross               & --               & --        & --       & \textbf{22.64}      & \textbf{0.855}      & \textbf{0.200}       \\
\midrule
\multirow{7}{*}{\rotatebox[origin=c]{90}{ {\scriptsize Color only}}}
& FVS~\cite{riegler2020free}                         &\redcross & --                & --        & --       & 20.00      & 0.790       & \underline{0.193}     \\
& NeRF~\cite{mildenhall2020nerf}                                &\redcross                & --               & --        & --       & 21.18      & 0.779      & 0.343     \\
& Mip-NeRF~\cite{barron2021mip}                             &\redcross                & --               & --        & --       & {21.54}      & {0.778}      & 0.365     \\
& PBNR~\cite{kopanas2021point}                         &\greencheck              & --                & --        & --       & 19.91      & 0.811      & {0.191}     \\
&PCL~\cite{liao2022kitti}                                 &\greencheck              & --                & --        & --       & 12.81      & 0.576      & 0.549     \\
& Gauss. Splat.~\cite{kerbl3Dgaussians}                  &\greencheck              & 18.59 &  0.642   & \textbf{0.257}   & \underline{22.08}          & \textbf{0.844}          & \textbf{0.139}         \\
\cmidrule[0.5pt](l{0em}r{0em}){2-9}
& PointNeRF~\cite{xu2022point}                           &\greencheck              & 17.63           & 0.629   & 0.337  & 19.44& {0.796} & 0.266 \\
& Ours                                &\greencheck              & \textbf{20.05}   & \textbf{0.665}   & \underline{0.305}  & \textbf{22.44} & \underline{0.828} & {0.212}   \\ 
\bottomrule
\end{tabular}
}
\vspace{-1.5em}
\end{table}

\begin{table}[t]
\caption{
{\bf Results on two ScanNet scenes~\cite{dai2017scannet} as pre-processed by \cite{xu2022point} --}
{Our method outperforms all others, especially \ky{PointNeRF, the method most similar to ours, by a large margin demonstrating the effectiveness of our multi-scale approach}}.
}
\resizebox{\linewidth}{!}{
\setlength{\tabcolsep}{8pt}
\begin{tabular}{lcccccc}
\toprule
                  &               & \multicolumn{3}{c}{Avg.} & \multicolumn{1}{c}{Scene-101} & \multicolumn{1}{c}{Scene-241} \\
                                                                 \cmidrule(lr){3-5} \cmidrule(lr){6-6} \cmidrule(lr){7-7}
                &\makecell{Uses\\points} & PSNR$\uparrow$    & SSIM $\uparrow$  & LPIPS$\downarrow$ & PSNR$\uparrow$        & PSNR$\uparrow$      \\
\midrule
NeRF~\cite{mildenhall2020nerf}              & \redcross   & 24.43   & 0.670      & 0.494 & 27.16        & 21.69           \\
Gauss. Splat.~\cite{kerbl3Dgaussians}  & \redcross   & {29.56}   & {0.812}    & {0.301} & {29.01}   & {30.11}       \\
Gauss. Splat.~\cite{kerbl3Dgaussians}  & \greencheck & \underline{29.93}   & \textbf{0.818}    & 0.275 & \underline{29.55}   & \underline{30.31}       \\
\midrule
PointNeRF~\cite{xu2022point}         & \greencheck & 25.92   & 0.784  & \underline{0.263} & 21.98    & 29.86       \\
Ours              & \greencheck & \textbf{30.56}   & \underline{0.808}  & \textbf{0.238} & \textbf{30.27}    & \textbf{30.85}       \\  
\bottomrule
\end{tabular}
}
\label{tab:scannet}
\vspace{-1.5em}
\end{table}

\begin{figure}
\centering
\includegraphics[width=\linewidth]{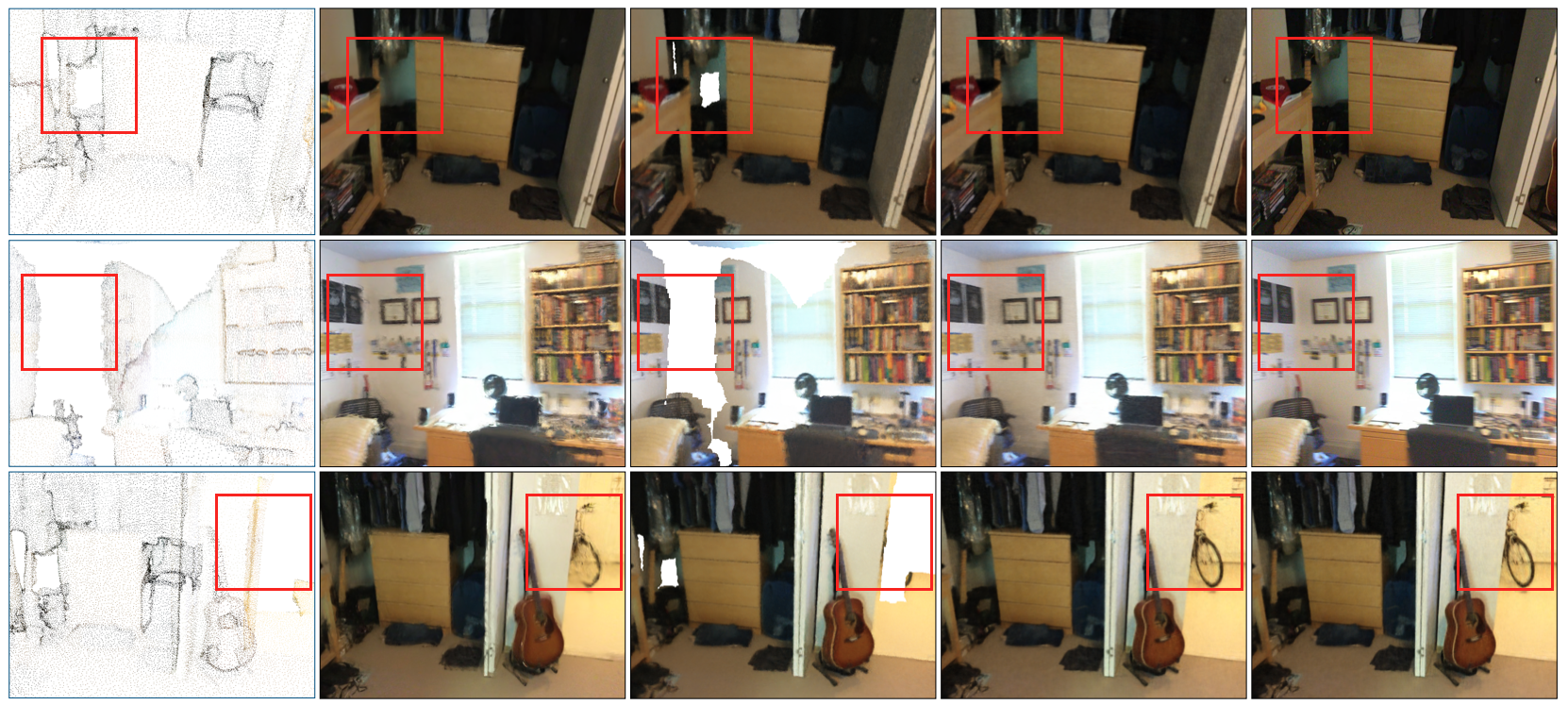}
\renewcommand{\arraystretch}{0.2}
\begin{tblr}{colspec={Q[2.0cm,c]X[c]X[c]X[c]X[c]}}
\tiny{(a) Input point cloud~\cite{dai2017scannet}} &
\tiny{(b) Gauss. Splat.~\cite{kerbl3Dgaussians}} &
\tiny{(c) PointNeRF~\cite{xu2022point}} & 
\tiny{(d) Ours} & 
\tiny{(e) Ground truth}\\
\end{tblr}
\caption{
{\bf Examples on \texttt{ScanNet} --}
PointNeRF fails to reconstruct the scene on regions where the point cloud is empty.
Both our method and Gaussian Splatting are able to fill them in, but our approach produces cleaner results, with fewer artifacts.
This is especially noticeable for Scene-101 (top and bottom rows), where the mesh has large holes where PointNeRF fails to render meaningful pixels, \ky{even with their `growing' heuristic that is aimed towards filling such gaps}.
}
\vspace{-1em}
\label{fig:qualitative_scannet}
\end{figure}

\subsection{ScanNet results -- \cref{fig:qualitative_scannet} and \cref{tab:scannet}}
Next, we consider indoor scans, using \texttt{ScanNet} \cite{dai2017scannet}. This dataset is less challenging than \texttt{KITTI-360}, but is a typical use-case for point cloud-based neural rendering, and is where the benefit of using point clouds was strongly demonstrated in PointNeRF~\cite{xu2022point}.

\paragraph{Baselines.}
We compare our method against NeRF~\cite{mildenhall2020nerf}, PointNeRF~\cite{xu2022point}, and Gaussian Splatting~\cite{kerbl3Dgaussians}.  
For the latter, we consider randomly initialized point clouds as well as those provided by the dataset.

\paragraph{Discussions.}
As shown in \cref{tab:scannet} and \cref{fig:qualitative_scannet}, our method performs best.
NeRF~\cite{mildenhall2020nerf}, for this dataset does not perform well as the scene is relatively textureless and smooth.
PointNeRF~\cite{xu2022point} improves over it by leveraging the point clouds.
It does, however, have issues on Scene-101, because its point cloud has large incomplete areas, which impair its performance, as shown in \cref{fig:qualitative_scannet}.
Our method is able to cope with these empty regions, thanks to our multi-scale framework.
Interestingly, Gaussian Splatting also works better than typical NeRF while trained purely with images, even when starting from random points---point cloud initialization can further improve its performance.
This suggests the importance of including the notion of locality introduced by points to the representation.
Our method outperforms all baselines, point-based or not.
We use point clouds from mesh inputs instead of depth images, also reported by PointNeRF, as the latter are \ky{extremely} dense (see~\cite[Tbl.~8]{xu2022point}).

\begin{table}[t]
\centering
\caption{
{\bf PSNR$\uparrow$ on NeRF Synthetic \cite{mildenhall2020nerf} --}
{Our method performs best overall, even on object-centric data with dense point clouds.}
}
\setlength\tabcolsep{2pt} %
\resizebox{\linewidth}{!}{
\begin{tabular}{@{}l c ccccccccc@{}}
\toprule
& \makecell{Uses\\points} & {\bf Avg.}   & Chair & Drums & Ficus & Hotdog & Lego  & Materials & Mic   & Ship  \\
\midrule
NeRF~\cite{mildenhall2020nerf}& \redcross &31.01 & 33.00 &25.01 &30.13 &36.18 &32.54 &29.62 &32.91 &28.65   \\
Plenoxels~\cite{yu_and_fridovichkeil2021plenoxels} & \redcross & 31.76    & 33.98 & 25.35 & 31.83 & 36.81  & 34.1  & 29.14     & 33.26 & 29.62 \\
InstantNGP~\cite{mueller2022instant} & \redcross & 33.18    & 35.00 & 26.02 & 33.51 & 37.40  & \underline{36.39} & 29.78     & {36.22} & {31.10}  \\
MipNeRF~\cite{barron2021mip} & \redcross & 33.09    & 35.14 & 25.48 & 33.29 & \underline{37.48}  & 35.7  & \textbf{30.71}     & \underline{36.51} & 30.41 \\
Gauss. Splat.~\cite{kerbl3Dgaussians} & \redcross  & {33.32}    & {35.83} & {26.15} & {34.87} & \underline{37.72}  & {35.78} & {30.00}     & {35.36} & {30.80}  \\
\midrule
FreqPCR~\cite{zhang2023frequency} & \greencheck  & 31.24    & 33.06 & 25.95 & 32.19 & 35.82  & 31.56 & 29.69     & 33.64 & 27.97 \\
TetraNeRF~\cite{kulhanek2023tetranerf} & \greencheck  & 32.53    & 35.05 & 25.01 & 33.31 & 36.16  & 34.75 & 29.30     & 35.49 & 31.13 \\
Gauss. Splat.~\cite{kerbl3Dgaussians} & \greencheck & \underline{33.60}	& \underline{36.10} &	\textbf{26.17} &	\textbf{34.87} &	\textbf{37.77} &	36.14&	30.15&	36.48&	\underline{31.12} \\
\midrule
PointNeRF~\cite{xu2022point} & \greencheck  & 31.77    & 35.09 & 25.01 & 33.24 & 35.49  & 32.65 & 26.97     & 35.54 & 30.18 \\
Ours   & \greencheck   & \textbf{33.70}    & \textbf{36.32} & \underline{26.11} & \underline{34.43} & {37.45}  & \textbf{36.75} & \underline{30.32}     & \textbf{36.85} & \textbf{31.34} \\ 
\bottomrule
\end{tabular}
}
\vspace{-1em}
\label{tab:nerfsyn}
\end{table}
\begin{figure}[t]
    \centering
    \includegraphics[width=.6\linewidth, trim={0 0 0 0},clip]{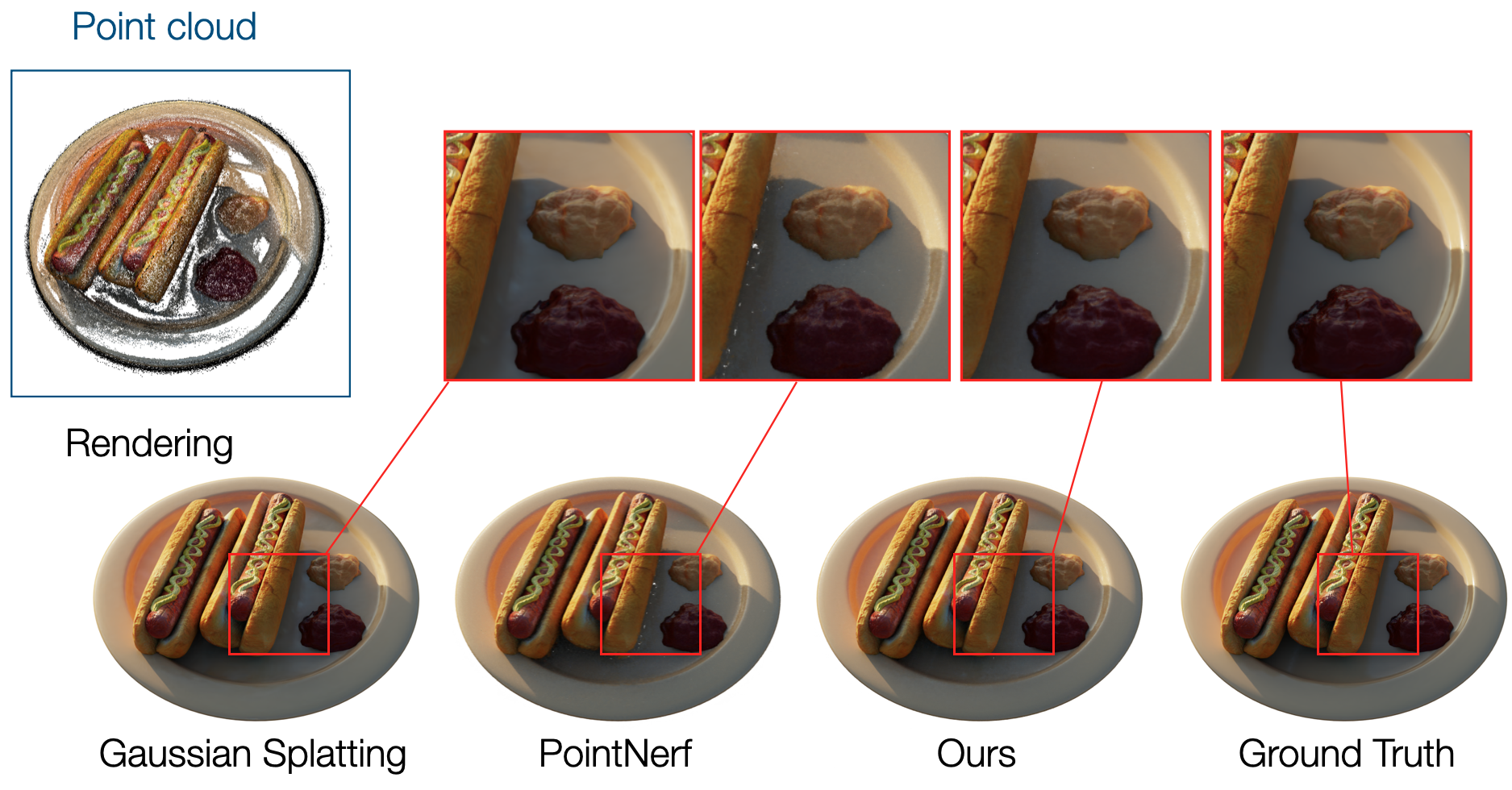}
    \caption{
    {\bf Examples on \texttt{NeRF Synthetic} --}
    Our multi-scale approach consistently fills in the holes in the input point cloud.
    PointNeRF relies on `pruning' and `growing' heuristics, which can fail where the point cloud is not sufficiently dense, as shown here.
    \ky{%
    While Gaussian Splatting also previous failed to fill in holes with their densification heuristics (\cref{fig:qualitative_kitti} and \cref{fig:qualitative_scannet}), for \texttt{NeRF Synthetic} these heuristics work well.
    }%
    } 
    \label{fig:qualitative_nerf}
    \vspace{-1em}
\end{figure}

\subsection{NeRF Synthetic results -- Fig.~\ref{fig:qualitative_nerf} and \cref{tab:nerfsyn}}

Finally, we verify the effectiveness of our method on \texttt{NeRF Synthetic}, to demonstrate that it remains helpful even \ky{use-cases designed for NeRF}.

\paragraph{Baselines.}
We compare our approach against both methods that only utilize RGB images~\cite{mildenhall2020nerf}, which this dataset is typically used to evaluate, and those that use point clouds~\cite{zhang2023frequency, kulhanek2023tetranerf, xu2022point}, including the recent Gaussian Splatting~\cite{kerbl3Dgaussians}.

\paragraph{Discussions.}
We report PSNR results in \cref{tab:nerfsyn} and provide qualitative examples in Fig.~\ref{fig:qualitative_nerf}.
Our method still performs best overall, slightly ahead of Gaussian Splatting.
More importantly, it outperforms the point-based baselines by a larger margin.

\subsection{Ablation study}
\label{sec:ablation}
We thoroughly ablate our method in this section on \texttt{KITTI-360} \ky{using the validation split}.
We consider the number of scale levels, using a tri-plane vs an MLP for $\featmlp$ at the coarsest scale levels, and study the effect of adding a global scale.
We also evaluate performance at different levels of point cloud sparsity.

\begin{table}[t]
\begin{minipage}[t]{0.56\linewidth}
    \centering
    \captionof{table}{
        {\bf Number of scales vs rendering quality --} As expected, more levels lead to better PSNR$\uparrow$. 
        We {\it always} use a global scale---`0' corresponds to using only a global scale, `1' the finest scale plus the global scale, and later adding coarser scale levels, up to our full model (`4').
    }
    \resizebox{\linewidth}{!}{
    \setlength{\tabcolsep}{18pt}
    \begin{tabular}{@{}lccc@{}}
    \toprule
    Num. of scales & PSNR$\uparrow$   & SSIM$\uparrow$  & LPIPS$\downarrow$ \\
    \midrule
    0 & 17.95 & 0.520 & 0.442 \\
    1 & 19.88 & \textbf{0.674} & \textbf{0.283} \\
    2 & 19.89 & 0.669 & 0.289 \\
    3 & 19.90 & 0.666 & 0.299 \\
    4 & \textbf{20.05} & {0.665} & 0.305 \\
    \bottomrule
    \end{tabular}
    }
    \label{tab:ablation_numscales} 
\end{minipage}
\hfill
\begin{minipage}[t]{0.42\linewidth}
    \centering
    \captionof{table}{
        {\bf Number of points --}
        We show that our method remains applicable to sparser point clouds, with noticeable improvement over points-agnostic model even at drastic downsampling rates ($1\%$).
    }
    \centering
    \resizebox{\linewidth}{!}{ 
    \setlength{\tabcolsep}{4pt}
    \begin{tabular}{@{}lccc@{}}
    \toprule
        Ratio of pts. & PSNR$\uparrow$   & SSIM$\uparrow$  & LPIPS$\downarrow$ \\
        \midrule
        0 & 17.95 & 0.520 & 0.442 \\  
        1 & 18.71	& 0.558	& 0.434 \\
        10 & 19.35	& 0.621	& 0.370 \\
        full & \textbf{20.05} & \textbf{0.665} & \textbf{0.305} \\
        \bottomrule
    \end{tabular}
    }
    \label{tab:ablation_numpts}
\end{minipage}
\end{table}
\begin{figure}[t]
    \centering
    \includegraphics[width=0.7\linewidth,trim={0 0.5cm 0 0},clip]{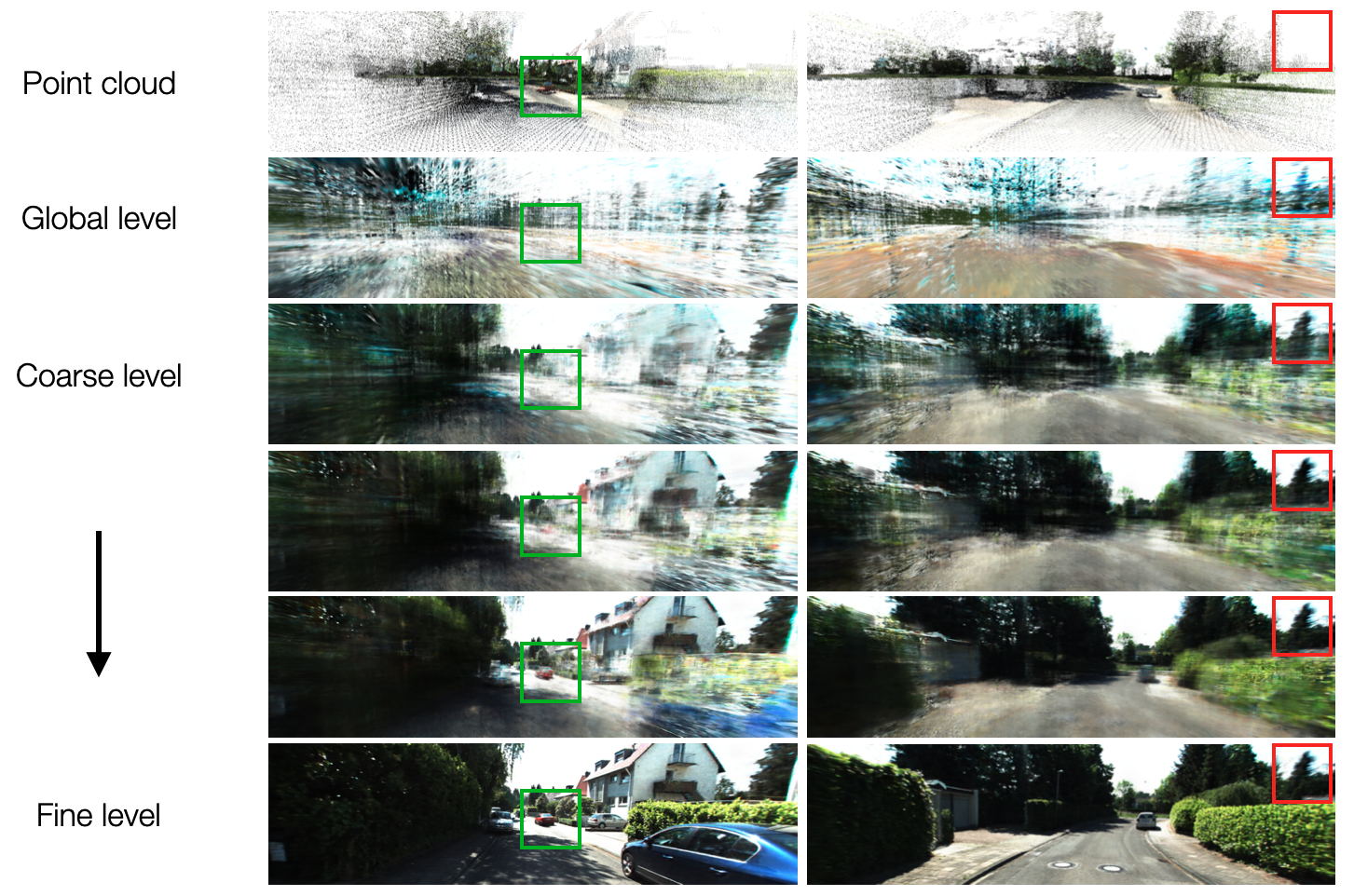}
    \caption{
    {\bf Rendering across scales --} 
    We study the behavior of our hierarchical approach by rendering an image adding one scale level at a time, from the global scale to the finest one. 
    As expected, the coarse scales are responsible for filling in empty regions (highlighted with red boxes) in the point cloud, different from the well-covered regions (highlighted with green boxes) that can be modeled via fine scales. 
    \vspace{-3mm}
    }
    \label{fig:ablation_numscale_example}
\end{figure}
\paragraph{Number of scale levels -- \cref{tab:ablation_numscales} and \cref{fig:ablation_numscale_example}}.
We ablate how the number of scale levels affects performance, by training and evaluating models using a different number of scales.
We also illustrate what each scale level is {\it adding}, by rendering views with a multi-scale model \ws{adding} one scale level at a time, in \cref{fig:ablation_numscale_example}. 
As clearly shown in the figure, the global scale is instrumental in rendering accurate pixels in those areas, and each successive scale adds finer details, improving the overall quality of the rendering.

\begin{table}[t]
\begin{minipage}[t]{0.4\linewidth}
    \centering
    \captionof{table}{
        {\bf Impact of the tri-plane --} 
        We evaluate tri-planes vs. MLPs. Using tri-planes for the coarsest scales improves performance, at a comparable computational cost.}
    \centering
    \resizebox{\linewidth}{!}{
    \setlength{\tabcolsep}{4pt}
    \begin{tabular}{@{}lccc@{}}
    \toprule
     & PSNR$\uparrow$   & SSIM$\uparrow$  & LPIPS$\downarrow$ \\
    \midrule
    MLP & 19.59	& 0.643	& 0.353 \\
    Triplane & \textbf{20.05} & \textbf{0.665} & \textbf{0.305} \\
    \bottomrule
    
    \end{tabular}
    }
    \label{tab:ablation_triplane} 
\end{minipage}
\hfill
\begin{minipage}[t]{0.56\linewidth}
    \centering
    \captionof{table}{
        {\bf Using a global feature --} 
        We measure PSNR$\uparrow$ on two datasets for different variants of our approach.
        Adding a global voxel at the coarsest scale to the hierarchical structure (right), improves performance (left), but is not sufficient by itself (middle).
        }
        \centering
        \resizebox{\linewidth}{!}{
        \setlength{\tabcolsep}{20pt}
        \begin{tabular}{@{}lccc@{}}
        \toprule
         & PSNR$\uparrow$   & SSIM$\uparrow$  & LPIPS$\downarrow$ \\
        \midrule
        w/o global & 19.78	& \textbf{0.675}	& \textbf{0.290} \\
        global only & 17.95	& 0.520	& 0.442 \\
        full & \textbf{20.05} & {0.665} & {0.305} \\
        \bottomrule
        \end{tabular}
        }
        \label{tab:ablation_localglobal} 
\end{minipage}
\end{table}
\paragraph{Tri-plane vs MLP -- \cref{tab:ablation_triplane}.}
We also provide an ablation study to evaluate the advantages of using a tri-plane instead of a regular MLP for the parameterized function $\featmlp$.
As outlined in \cref{sec:triplane}, we always use MLPs at the two finest scale levels. Using a tri-plane performs slightly better at a similar computational cost.

\paragraph{Using a global voxel -- \cref{tab:ablation_localglobal}.}
We evaluate the importance of adding a global voxel at the coarsest scale. We compare three variants of our method: one using four local scales (``w/o global''); one using only the global scale (``global-only''), \ie, a traditional point-agnostic NeRF; and one using both (``full''). 
The point-based formulation outperforms point-agnostic NeRF, but combining them with our multi-scale-plus-global approach does best. 

\paragraph{Number of points -- \cref{tab:ablation_numpts}.}
We measure performance while randomly downsampling the point cloud with increasing ratios. 
Our approach performs while even at downsampling rates below $1\%$ and using as few as \ws{10k} points.

\section{Conclusions}
\label{sec:conclusions}

Neural Radiance Fields are a paradigm shift in novel-view synthesis. Despite their promise, and a large number of follow-up papers, challenges persist, particularly when few views of the scene are available. Point clouds provide a very attractive data stream, complementary to images, and often readily available in both indoor and outdoor settings---but have a different set of challenges, due to incompleteness and sparsity.
We mitigate this with a simple yet novel multi-scale representation that combines global and local information, yielding significant performance improvements across the board. Our work unifies point cloud-based and standard NeRF pipelines and adapts effectively to variable point densities and empty regions, pushing novel view synthesis on uncontrolled, real-world data closer to practice.

\paragraph{Limitations and future work.} 
\ky{%
While our method provides significant improvements over PointNeRF by combining point-based ones with classic NeRF, the computational cost of our method is naturally bound by classic NeRF---on \texttt{NeRF Synthetic}, our method induces a 20\% in compute overhead to the classic NeRF backbone that we use.
}%
Thus, an interesting research direction would be to combine the strengths of our multi-scale strategy in handling incomplete and sparse point clouds
\ky{%
with the high computational-efficiency of 3D Gaussian Splatting~\cite{kerbl3Dgaussians}, especially given the pitfalls of 3D Gaussian Splatting demonstrated in our work.
}%

\bibliographystyle{splncs04}
\bibliography{main}

\appendix   
\clearpage
\setcounter{page}{1}

   {
   \newpage
        \centering
        \Large
        \vspace{0.5em}Supplementary Material \\
        \vspace{1.0em}
   }

We detail mutl-scale point generation and coordinate system of the global triplane.  
Furthermore, in the provided 
\url{https://pointnerfpp.github.io}, 
we provide more rendering results.

\section{Multi-scale Point Generation via Grid-subsampling}
We build multi-scale points from an input point cloud using grid subsampling, which is more robust to varying density as shown in KPConv~\cite{thomas2019kpconv}.
Specifically, a new support point at each scale is the barycenter of the original input points contained in a grid cell. 
Thereby, we control the scale and density at each level via the grid size of cells. 

We set the grid size at the level s as $\omega * \gamma^{s-1}$ where $\omega$ is the initial grid size at the first level and $\gamma$ is the stride size.
We use the larger grid size for severely incomplete point clouds and a small grid size for the complete point cloud.  
Specifically, for \texttt{KITTI-360} ~\cite{liao2022kitti}, we set $\omega$ as 8cm and $\gamma$ as 2.92. As a result, the grid size at the coarsest point level (i.e., s=4) is 2 meters.  
For \texttt{ScanNet}~\cite{dai2017scannet}, we set $\omega$ as 0.008 and $\gamma$ as 2.0. 
For \texttt{Nerf Synthetic}~\cite{mildenhall2020nerf} where point clouds are relatively complete, we set $\omega$ as 0.004 and $\gamma$ as 1.6.

\section{The Coordinate System of Global Triplane}
We align world coordinate system and the normalized coordinate system of global triplane. 
We utilize the principal component analysis (PCA) to calculate the reference coordinate frame of input point cloud.
The resultant reference frame consists of rotation, translation and scale, thereby defining the alignment matrix transforming world coordinates to coordinate system of global triplane.
In \texttt{ScanNet}~\cite{dai2017scannet} and \texttt{Nerf Synthetic}~\cite{mildenhall2020nerf} where points distribute uniformly along three axes and center at the origin, we simply normalize world coordinates using the scale part of reference frame.
For \texttt{KITTI 360}~\cite{liao2022kitti}, we use full reference frame instead -- i.e., we rotate, translate and scale the world coordinates -- because, in this dataset, the car moves along one major direction, leading to the points heavily unbalanced along three axes.
With this PCA-based canonicalization, we compactly compress all possible query points into triplane's normalized frame, allowing for fully utilizing the capacity of global triplane.
\section{More rendering results}
We furthermore provide more rendering results -- rendering more frames and forming videos. For more details, please refer to
\url{https://pointnerfpp.github.io}. 
\end{document}